\pdfoutput=1

\documentclass[11pt]{article}

\usepackage[preprint]{acl}

\usepackage{times}
\usepackage{latexsym}

\usepackage[T1]{fontenc}

\usepackage[utf8]{inputenc}

\usepackage{microtype}
\usepackage{inconsolata}

\usepackage{mathtools}
\usepackage{graphicx}
\usepackage{custom_style}
\usepackage{enumitem}

\usepackage{caption}
\usepackage{subcaption}
\usepackage{pgfplotstable}
\usepackage{tabularx}
\usepackage{multirow}
\usepackage{multicol}
\usepackage{booktabs}
\title{Task-driven Layerwise Additive Activation Intervention}

\author{
  Hieu Trung Nguyen$^1$ \hspace{15pt} Bao Nguyen$^1$ \hspace{15pt} Binh Nguyen$^2$ \hspace{15pt} Viet Anh Nguyen$^1$ \\[2mm]
  $^1$ The Chinese University of Hong Kong \\
  $^2$ National University of Singapore \\
  \texttt{thnguyen@se.cuhk.edu.hk,nbnguyen@se.cuhk.edu.hk} \\
  \texttt{binhnt@nus.edu.sg,nguyen@se.cuhk.edu.hk}
}

\usepackage{amsmath,amssymb,amsthm}
\usepackage{bm}
\usepackage{algorithm,algorithmic}

\newcommand{\R}{\mathbb{R}}

\begin{document}
\maketitle
\begin{abstract}
Modern language models (LMs) have significantly advanced generative modeling in natural language processing (NLP). Despite their success, LMs often struggle with adaptation to new contexts in real-time applications. A promising approach to task adaptation is activation intervention, which steers the LMs' generation process by identifying and manipulating the activations. However, existing interventions are highly dependent on heuristic rules or require many prompt inputs to determine effective interventions. This paper proposes a layer-wise additive activation intervention framework that optimizes the intervention process, thus enhancing the sample efficiency. We benchmark our framework on various datasets, demonstrating improvements in the accuracy of pre-trained LMs and competing intervention baselines.
\end{abstract}

\section{Introduction}

Transformer-based language models (LMs) have revolutionized generative modeling for natural language processing (NLP). This is demonstrated by the impressive performances of LMs in various important NLP tasks \citep{ref:radford2019language,ref:brown2020language,ref:achiam2023gpt,ref:touvron2023llama,ref:jiang2023mistral,ref:abdin2024phi,ref:anthropic2024claude,ref:dubey2024llama}. One of such is in-context learning (ICL, \citealt{ref:brown2020language}), where a pretrained LM can perform NLP tasks without fine-tuning their parameters. This is achieved by providing the model with prompts that include demonstrations of the task, allowing it to learn from the examples and make predictions without requiring additional training. Despite this, performing ICL on LMs remains challenging, as LMs still struggle to adapt quickly to new context shifts in real-time applications.  

One possible method for adaptation is \emph{activation intervention} \citep{ref:subramani2022extracting,ref:turner2023activation,ref:hernandez2023measuring,ref:todd2023function,ref:li2024context,nguyen2025risk,jiang2025probefreelowrankactivationintervention}, where one uses the activations of the model that are most likely responsible for ICL to steer the generation process. However, most of these works either derive the intervention based on a heuristic rule or require a large amount of prompt input. 

\textbf{Contributions.} In this work, we aim to design a principled, optimization-based intervention that delivers competitive results with limited training demonstrations. We propose a layerwise additive activation intervention method for task-driven learning. The intervention is an optimal vector that minimizes the mismatch between the intervened decoding output and the target desired output in the training data. Additionally, we impose a joint lasso and group lasso regularization to mitigate overfitting on the sample size and promote the component and head sparsity of the intervention. 

Existing activation intervention methods scatter the interventions across multiple layers~\cite{ ref:todd2023function,ref:turner2023activation,ref:li2024inference}, which can negatively affect the effectiveness of the intervention at later layers due to the representation shifts of the activations generated at earlier layers. To address this issue, we propose to focus the intervention on the same layer, which can be easily formulated as a layerwise optimization problem. The layerwise optimization problem has shown effectiveness in driving the LLM-generated content to human alignment~\cite{nguyen2025risk,jiang2025probefreelowrankactivationintervention}. Moreover, our intervention can facilitate task calculus by focusing on the same layer across tasks. By an additive composition of different task-specific interventions, we obtain a new intervention for the corresponding composition of tasks, as we will demonstrate in the numerical experiments.

\section{Related Works}

\textbf{In-Context Learning.} Since its introduction by \citet{ref:brown2020language}, ICL in LM has been studied extensively in various directions. For example, \citet{ref:reynolds2021prompt,ref:yoo2022ground} analyzed the role of prompts in improving the ICL performance. Theoretical analysis of how LMs perform ICL has been proposed by \citet{ref:akyurek2022learning,ref:dai2023can,ref:von2023transformers,ref:sander2024transformers}. These works study the internal mechanism -- either with regularized linear regression or gradient descent -- of the transformer architecture, which is the workhorse behind most current state-of-the-art LMs. 

\textbf{Language model intervention.} Intervening on the hidden states of transformer-based LMs, or activations editing, has recently emerged as an efficient method for controllable text generation. Contrasting to weights editing, activations editing refers to modifying the output of attention heads on one or several layer(s) of the transformer architecture, ultimately steering the generated text to desirable outcomes. Initially proposed to perform text style transfer, this method has been extended to improve the performance of few shots / zero shots of ICL, such as in \citet{ref:todd2023function,ref:liu2023context,ref:hendel2023incontext,ref:li2024context,ref:hernandez2024linearity}. Our work follows this direction but improved upon them by using only a fewer number of prompt inputs. As such, the aforementioned works, most notably by \citet{ref:todd2023function}, are directly related to our work.

\section{Methodologies} \label{sec:intervention}

We have a pre-trained decoder-only transformer-based LM (for example, LLama3-8b) that is not yet fine-tuned for the few-shot in-context learning task (ICL). The LM has $L$ layers; each layer has $H$ heads of dimension $d$; overall, the activation vector at each layer has a dimension $D = d\times H$. We use $\ell \in \{1, \ldots, L\}$ as the layer index, and use $h \in \{1, \ldots, H\}$ as the head index. For Llama3-8b, we have $L = 32$, $H = 32$ and $d = 128$.

We consider the layer-wise intervention consisting of finding a task-specific modification vector to be added to the activations of the input's last token so that the LM's output is steered toward our desired direction. To formalize this problem, we consider a task $\tau$ dataset consisting of $N_\tau$ samples. Each sample $i$, $i = 1, \ldots, N_\tau$, can be described by a tuple $(s_{i\tau}, r_\tau, t_{i\tau})$, where $s_{i \tau}$ is the input text, $r_\tau$ is a special token padded to the end of the input, and $t_{i\tau}$ is the desired (ground-truth) target output corresponding to the input $s_{i\tau}$. When there is no possible confusion, we will omit the task index $\tau$ to avoid cluttered notation.

Our method aims to find a task-specific $\Delta$ from the training data. Then, at inference time with a test input $s_{\mathrm{test}}$, we intervene by adding $\Delta$ to the activations of the last token corresponding to the input $(s_{\mathrm{test}}, r)$ to generate $\hat{t}_{\mathrm{test}}$. The success of the intervention is measured by the discrepancy in the test set between the generated output $\hat{t}_{\mathrm{test}}$ and the true desired output $t_{\mathrm{test}}$.

The last token's activations at layer $\ell$ of the input $(s_i, r)$ are denoted by $a_\ell(s_i, r)$; consequently, the additively-intervened activations become $a_\ell(s_i, r) + \Delta$. The activations at the last layer (layer $L$) after the intervention become $a_{L, \Delta}(s_i, r)$. The decoder will transform $a_{L, \Delta}(s_i, r)$ into the distribution of the next token for generation. A good intervention vector $\Delta$ should minimize the generation loss averaged over the training dataset
    \begin{equation} \label{eq:loss}
        \mathrm{Loss}(\Delta) = \frac{1}{N} \sum_{i=1}^N \mathcal L_{\mathrm{task}}( a_{L, \Delta}(s_i, r), t_i).
    \end{equation}
Simply minimizing~\eqref{eq:loss} leads to overfitting: in general, the number of training samples $N$ is small, while the dimension $D$ of the vector $\Delta$ is much larger ($D = 4096$ for Llama3-8b). We propose using both lasso regularization and group lasso regularization to combat overfitting. Thus, the intervention $\Delta$ solves
\begin{equation} \label{eq:problem}
    \min_{\Delta \in \R^D}~ \mathrm{Loss}(\Delta) + \gamma \| \Delta \|_1 +  \lambda \sum_{h=1}^H \| \Delta_h \|_2,
\end{equation}
where $\gamma > 0$ is a lasso parameter controlling the sparsity of $\Delta$, and $\lambda > 0$ is a group lasso parameter. Here, a natural group assignment is by head, where we decompose $\Delta = (\Delta_1, \ldots, \Delta_H)$, where each $\Delta_h \in \R^d$. The group lasso term penalizes the sum of the 2-norm of headwise interventions $\Delta_h$. We choose group lasso regularization to promote sparsity \emph{within heads of activations}, as empirical evidence from previous work such as \citet{ref:hernandez2023inspecting,ref:todd2023function} and \citet{ref:li2024inference} suggests that only a portion of attention heads is responsible for the transformer's ability to generate controllable outputs. The lasso penalty is also added to promote an additional degree of sparsity across all elements of $\Delta$.  

Next, we describe two specific applications of this task-driven intervention.

\begin{table*}[!ht]
\centering
\caption{Results for single rule understanding task. Our optimization-based method outperforms the baselines in both metrics.}
\label{tab:rule}
\tiny
\begin{filecontents*}{single_rule_understanding.csv}
Method,English-French,English-French,English-German,English-German,Antonym,Antonym,Synonym,Synonym
,Exact Match $\uparrow$,GPT-Eval $\uparrow$,Exact Match $\uparrow$,GPT-Eval $\uparrow$,Exact Match $\uparrow$,GPT-Eval $\uparrow$,Exact Match $\uparrow$,GPT-Eval $\uparrow$
0-shot prompting,0.069 $\pm$ 0.012,0.02,0.022 $\pm$ 0.005,0.04,0.050 $\pm$ 0.026,0.09,0.051 $\pm$ 0.012,0.115
10-shot prompting,0.000 $\pm$ 0.000,0.188,0.075 $\pm$ 0.014,0.03,0.000 $\pm$ 0.000,0.129,0.000 $\pm$ 0.000,0.109
0-shot prompting FV,0.129 $\pm$ 0.042,0.173,0.054 $\pm$ 0.012,0.08,0.000 $\pm$ 0.000,0.099,0.124 $\pm$ 0.023,0.179
10-shot prompting FV,0.241 $\pm$ 0.053,0.267,0.123 $\pm$ 0.031,0.133,0.056 $\pm$ 0.013,0.178,0.122 $\pm$ 0.028,0.614
\textbf{Ours},\textbf{0.795 $\pm$ 0.024},\textbf{0.768},\textbf{0.620 $\pm$ 0.041},\textbf{0.872},\textbf{0.514 $\pm$ 0.063},\textbf{0.902},\textbf{0.349 $\pm$ 0.085},\textbf{0.74}
\end{filecontents*}
\pgfplotstabletypeset[
    col sep=comma,
    string type,
    every head row/.style={before row=\toprule},
    every row no 0/.style={after row=\midrule},
    every head row/.style={output empty row, before row={%
            \toprule \multirow{2}{*}{Method}  &
            \multicolumn{2}{c}{Eng-Fr} & \multicolumn{2}{c}{Eng-Ger} & \multicolumn{2}{c}{Antonym} & \multicolumn{2}{c}{Synonym}  \\
            \cmidrule(r){2-3} \cmidrule(r){4-5} \cmidrule(r){6-7} \cmidrule(r){8-9}
        }},
    every last row/.style={after row=\bottomrule},
]{single_rule_understanding.csv}
\vspace{-3mm}
\end{table*}

\subsection{Rule Understanding}

The first application of the layer-wise task-specific activation is the rule understanding task~\cite{ref:todd2023function, ref:hernandez2024linearity}. Each sample consists of a tuple (subject, relation, object), equivalently denoted by $(s_i, r, o_i)$, where $s_i$ is a phrase, $r$ is the special relationship token, and $o_i$ is the output. For example, an exemplary sample is of the form~\texttt{hello:bonjour}, where \texttt{hello} is $s_i$, \texttt{:} is the special token $r$, and \texttt{bonjour} is $o_i$. This particular sample is picked from the task of translating an English phrase into French, which a knowledgeable human can easily deduce. Nevertheless, this conceptual description of the task is not given to the model. The goal of the intervention vector $\Delta$ is to steer the LM to generate the corresponding French translation of the input word. 

In this problem, the target $t_i$ is the next token $o_i$ in the training data. An effective loss here is the negative log-probability of the token $o_i$ from the decoder: if the decoder outputs a distribution over the dictionary $\mathrm{DEC}(a_{L, \Delta}(s_i, r))$, then,
\begin{align*} \notag
    &\mathcal L_{\mathrm{task}}(a_{L, \Delta}(s_i, r), o_i) \\
    &\qquad =- \log \mathrm{DEC}(a_{L, \Delta}(s_i, r))[o_i].
\end{align*}

\subsection{Opinion Generations}

The second application we consider is the opinion elicitation problem~\cite{ref:santurkar2023whose}, where the whole population consists of multiple groups. Each group has its own characteristics, leading to a different group-specific distribution of responses to the input question. In this problem, each group is considered as one task; the training datasets consists of multiple textual questions $s_i$, padded with the special token $r$, and the response distribution is $\pi_i$ supported on the target response alphabet $\mathcal O_i$. 

Here, we set the target $t_i$ as the distribution $\pi_i$, and the task loss is the Kullback-Leibler divergence between the decoding distributions over the response alphabet $\mathcal O_i$ and the target $\pi_i$:
\begin{align*}
     &\mathcal L_{\mathrm{task}}(a_{L, \Delta}(s_i, r), \pi_i) \\
     &\qquad = \mathrm{KL}( \mathrm{DEC}(a_{L, \Delta}(s_i, r))[\mathcal O_i] \parallel \pi_i).
\end{align*}

\section{Numerical Experiments}

We perform benchmarks to demonstrate our algorithm's performance on two tasks: Rule Understanding and Opinion Dynamics. All experiments are run on 4$\times$ NVIDIA A5000 GPUs. Our implementation will be published at \url{https://github.com/HieuNT91/LayerwiseIntervention.git}

\begin{figure}
    \centering
    \includegraphics[width=0.4\textwidth]{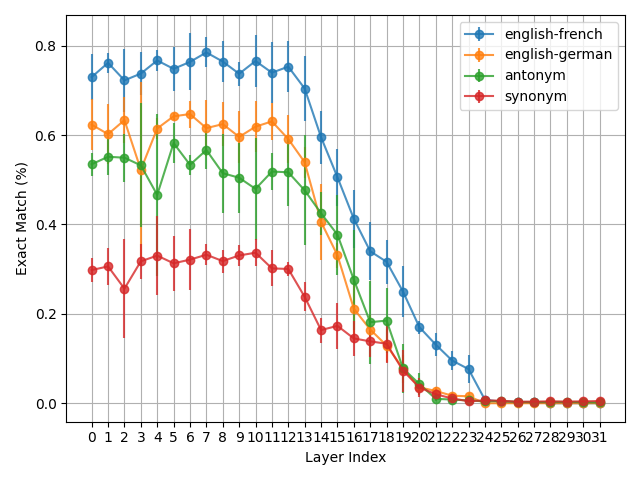}
    \caption{Average Exact Match for unregularized interventions at different layers. Results are averaged over five random seeds.}
    \label{fig:layer_idx}
\end{figure}

\begin{table*}[!h]
\centering
\small
\caption{Results for composition rule understanding task. Re-optimizing the intervention vectors delivered better results, but the addition of the task vector (first row) without optimization still shows comparatively good performance.}
\label{tab:ablation}
\scriptsize
\begin{filecontents*}{composite_rule_understanding.csv}
Method,English to French Antonym,English to French Antonym,English to German Antonym,English to German Antonym,English to French Synonym,English to French Synonym,English to German Synonym,English to German Synonym
,Exact Match $\uparrow$,GPT-Eval $\uparrow$,Exact Match $\uparrow$,GPT-Eval $\uparrow$,Exact Match $\uparrow$,GPT-Eval $\uparrow$,Exact Match $\uparrow$,GPT-Eval $\uparrow$
Ours (Add),0.324 $\pm$ 0.140,0.731,0.237 $\pm$ 0.046,0.312,0.644 $\pm$ 0.125,0.852,0.601 $\pm$ 0.215,0.901
Ours (Re-optimized),0.551 $\pm$ 0.076,0.896,0.546 $\pm$ 0.053,0.724,0.768 $\pm$ 0.036,0.937,0.780 $\pm$ 0.046,0.984
\end{filecontents*}
\pgfplotstabletypeset[
    col sep=comma,
    string type,
    every head row/.style={before row=\toprule},
    every row no 0/.style={after row=\midrule},
    every head row/.style={output empty row, before row={%
            \toprule \multirow{2}{*}{Method}  &
            \multicolumn{2}{c}{Eng-Fr Antonym} & \multicolumn{2}{c}{Eng-Ger Antonym} & \multicolumn{2}{c}{Eng-Fr Synonym} & \multicolumn{2}{c}{Eng-Ger Synonym} \\
            \cmidrule(r){2-3} \cmidrule(r){4-5} \cmidrule(r){6-7} \cmidrule(r){8-9}
        }},
    every last row/.style={after row=\bottomrule},
]{composite_rule_understanding.csv}
\vspace{-3mm}
\end{table*}

\begin{table*}[!h]
    \centering
    \caption{Kullback-Leibler mismatch for the opinion dynamic task using OpinionQA dataset with different subgroups of the population. Smaller values are better.} 
    \small
    \label{tab:opinionQA}
    \begin{tabular}{cccccc}
        \toprule
        Method & 100,000 USD or more & Less than 30,000 USD & Moderate & Northeast & Average\\
        \midrule
        0-shot Prompting & 2.761 & 2.451 & 3.451 & 4.131 & 3.200\\
        10-shot Prompting & 1.665 & 2.047 & 2.342 & 2.244 & 2.074\\
        \textbf{Ours} & \textbf{0.283} & \textbf{0.260} & 
        \textbf{0.260} & \textbf{0.288} & \textbf{0.273}\\
        \bottomrule
    \end{tabular}
\end{table*}

\subsection{Single Rule Understanding}
We utilize four tasks from \citet{ref:todd2023function}: Antonym, Synonym, English-French, and English-German; the task description is relegated to the Appendix~\ref{sec:datasets}. We select these tasks because the empirical results from~\citet{ref:todd2023function} indicated that the non-optimization interventions perform poorly on these tasks. We can access $N = 10$ pairs of input and output samples for each dataset and intervene at layer $\ell = 4$.

We use two performance metrics:
\begin{itemize}[leftmargin=5mm]
    \item Exact Match: the proportion of predictions that match exactly the targets. 
    \item GPT-Eval measures the proportion of predictions confirmed true for a task by GPT-4. An input can lead to multiple reasonable outputs in almost all tasks. For example, an English word can have multiple synonyms. Therefore, we design a specific query format for each task to ask GPT-4, the state-of-the-art large language model, to confirm the answer. Detailed information on the query format for each task is provided in the appendix. To minimize uncertainty in GPT-4's responses, we query GPT-4 five times for each input-prediction pair. The prediction is deemed acceptable if GPT-4 confirms the prediction as suitable for the input in more than two out of the five attempts.
\end{itemize}

We compare our interventions against four baselines: (i-ii) zero- and ten-shot prompting, (iii-iv) zero- and ten-shot prompting using the function vector (FV) method proposed in~\citet{ref:todd2023function}. The results in Table~\ref{tab:rule} show a significant improvement in rule understanding across multiple tasks using our proposed method compared to the baselines. The performance gains are also consistently shown in semantic relationship tasks (antonyms and synonyms). Notably, the performance gaps are large compared with zero-shot and few-shot prompting baselines (with and without adding Function Vectors). The main reason for the performance difference is that our method is based on a smaller training sample size, and task signals are efficiently extracted in the optimization process.

\subsection{Rule Understanding Composition}

Tasks can be easily composed: if $\tau$ is the antonym task and $\tau'$ is the English-French translation task, then one can compose $\tau' \circ \tau$ that takes an English word as input and generates the corresponding French-antonym as output. In this section, we test the algebraic additive composition of the trained intervention vectors. We assume that we have two intervention vectors at the same layer $\ell$ denoted as $\Delta_{\tau}$ and $\Delta_{\tau'}$ for the task $\tau$ and task $\tau'$, respectively.
We define a simple algebra sum between these two interventions to form a new one $\Delta_{\tau,\tau'} = \Delta_{\tau} + \Delta_{\tau'}$. Next, we study whether the new vector $\Delta_{\tau,\tau'}$ can be used for the composition task $\tau' \circ \tau$. We expect $\Delta_{\tau, \tau'}$ to perform competitively on the newly composed task.

In Table~\ref{tab:ablation}, we present the results obtained by two methods: (i) by adding intervention vectors as previously described and (ii) by re-optimizing the interventions on the composed tasks' training data (using 10 training samples). Clearly, we expect that re-optimizing will deliver better results, as reflected in Table~\ref{tab:ablation}. Nevertheless, we observe that the performance of the additive composition remains competitive.

\vspace{-5pt}
\subsection{Opinion Dynamic}

We use the OpinionQA dataset~\cite{ref:santurkar2023whose, ref:zhao2023group}, which evaluates how closely language models align with the opinions of certain groups in the whole population. We use zero-shot and ten-shot prompting as the baselines. Further, we use the Kullback-Leibler divergence between language models' opinion distribution and human distribution as a performance metric. We report the results on the test set in Table~\ref{tab:opinionQA}. Our method outperforms the prompting baselines and better matches the group-specific distributions.

\subsection{Additional Ablation Studies}
We conduct multiple ablation studies to validate our design choices and demonstrate the versatility of our approach.

\subsubsection{Regularized vs. Unregularized Loss}
To assess the contribution of the regularization terms in our loss function~\eqref{eq:problem}, we compare the performance of models trained with and without regularization ($\lambda = \gamma = 0.01$ vs. $\lambda = \gamma = 0$). Table~\ref{tab:ablated_loss} shows that incorporating the regularization term improves performance across all tasks, especially on the translation tasks.

\begin{table}[t]
    \centering
    \small
    \caption{Performance comparison between the unregularized and regularized loss on four tasks. We use Exact Match to measure performance on each task.  Higher values are better.} \label{tab:ablated_loss}
    \begin{tabular}{p{1.6cm}cp{1.2cm}p{1cm}c}
        \toprule
        Method & Eng-Fr & Eng-Ger  & Antonym& Synonym\\
        \midrule
        Unregularized & 0.504 & 0.302 & 0.371 & 0.314 \\
        \textbf{Regularized} & \textbf{0.795} & \textbf{0.620} & \textbf{0.514}	& \textbf{0.349} \\
        \bottomrule
    \end{tabular}
\end{table}

\subsubsection{Experiments with Other Language Models}
To demonstrate the generalizability of our approach across different architectures and model sizes, we experimented with three language models: Mistral-7B-v0.3~\cite{ref:jiang2023mistral}, Gemma2-2B~\cite{ref:team2024gemma}, and Llama3-8B~\cite{ref:touvron2023llama}. Table~\ref{tab:ablated_models} summarizes the performance on the Eng-Fr, Eng-Ger, and Antonym tasks. Notably, Llama3-8B achieves the best overall performance, indicating that our method scales favorably with increased model capacity.

\begin{table}[h]
    \centering
    \small
    \caption{Performance of various language models on selected tasks. We use Exact Match to measure performance on each task. Higher values are better.} \label{tab:ablated_models}
    \begin{tabular}{cccc}
        \toprule
        Model & Eng-Fr  & Eng-Ger& Antonym \\
        \midrule
        Mistral-7B-v0.3 & 0.521 & 0.385 & 0.321 \\
        Gemma2-2B & 0.710 &	0.221 & 0.314 \\
        \textbf{Llama3-8B} & \textbf{0.795} & \textbf{0.620} & \textbf{0.514} \\
        \bottomrule
    \end{tabular}
\end{table}

\subsubsection{Comparison with Intervention and Finetuning Baselines.}
We compare our approach with three fine-tuning baselines using the standard implementation provided by the PEFT library~\cite{ref:mangrulkar2022peft} and one intervention baseline using author implementation\footnote{\url{https://github.com/shengliu66/ICV.git}}. This comparison evaluates the effectiveness of our method in the low-sample size settings. Below, we briefly describe each baseline:
\begin{itemize}[leftmargin=5mm]
    \item In-Context Vector (ICV)~\citep{ref:liu2023context}: To imitate the 10-shot setting, we use 10 examples and the default step size of 0.1 to generate the in-context vector.
    \item $\text{IA}^3$~\citep{ref:liu2022few}: we applied adapters to the $\texttt{k}_\texttt{proj}$, $\texttt{v}_\texttt{proj}$  and $\texttt{down}_\texttt{proj}$ layers of the network. Specifically, the $\text{IA}^3$ vectors were multiplied with the input to the $\texttt{down}_\texttt{proj}$ layer to scale the activations accordingly.
    \item Soft Prompt~\citep{ref:lester2021power}: We initialized the first token with the task description, e.g., `the French translation of this word', and fine-tuned eight additional virtual tokens with this initial prompt.
    \item LoRA~\citep{ref:hu2021lora}: We fine-tuned a rank-4 matrix, introducing an additional 53,248 parameters to the model.
\end{itemize}
Table~\ref{tab:finetuning} summarizes the performance of these baselines on a Rule Understanding task. Our method consistently outperforms the baseline approaches across multiple tasks, demonstrating its robustness in low-data scenarios.
\begin{table}[h]
    \centering
    \small
    \caption{Comparison with intervention baseline and finetuning baselines. We use Exact Match to measure performance on each task. Higher values are better.} \label{tab:finetuning}
    \begin{tabular}{cccc}
        \toprule
        Method & Eng-Fr  & Eng-Ger  & Antonym \\
        \midrule
        ICV & 0.396	& 0.423 & 0.008 \\
        $\text{IA}^3$ & 0.521 & 0.385 & 0.321 \\
        Soft Prompt & 0.710 &	0.221 & 0.314 \\
        LoRA & 0.681 & 0.606 &	0.427 \\
        \textbf{Ours} & \textbf{0.795} & \textbf{0.620} & \textbf{0.514} \\
        \bottomrule
    \end{tabular}
\end{table}

\section{Conclusions}
In this paper, we propose and showcase an effective approach using layer-wise additive activation interventions to steer the output of LMs. Our approach effectively enhances the model performance by optimizing an intervention vector to minimize the mismatch between the intervened decoding output and the desired target output in the training data. Additionally, incorporating both lasso and group lasso regularizations addresses overfitting and promotes sparsity in activation heads, ensuring efficient interventions. Our evaluations on the rule understanding task and the opinion dynamic task demonstrate that this method significantly improves the performance of pre-trained LMs across various tasks, outperforming existing intervention techniques.

\section{Limitations}

The main limitation of our approach is that we require access to the model's activations. However, this limitation is relevant for \textit{any} activation intervention method in the literature, including \citet{ref:li2024inference} and \citet{ref:todd2023function}, due to the nature of the approach. In this paper, we have shown that our interventions are effective in the Llama3-8b model, and we expect that the intervention will also be effective in larger models such as Llama3-70b. 

Although we use interventions to steer the output to adapt to tasks, it is foreseeable that these techniques can be used for possibly unethical purposes, such as generating untruthful or toxic texts. Thus, we strongly recommend studying possible defenses for these problems.

\noindent\textbf{Acknowledgments.} Viet Anh Nguyen gratefully acknowledges the generous support from the UGC Early Career Scheme Grant 24210924 and the CUHK’s Improvement on Competitiveness in Hiring New Faculties Funding Scheme. Binh Nguyen is supported by NUS Start-up Grant A-0004595-00-00.

\bibliography{bib}
\newpage
\appendix

\section{Effects of Regularization}

\begin{figure}[h]
    \centering
     \begin{subfigure}[b]{0.52\textwidth}
         \centering
         \includegraphics[width=\textwidth]{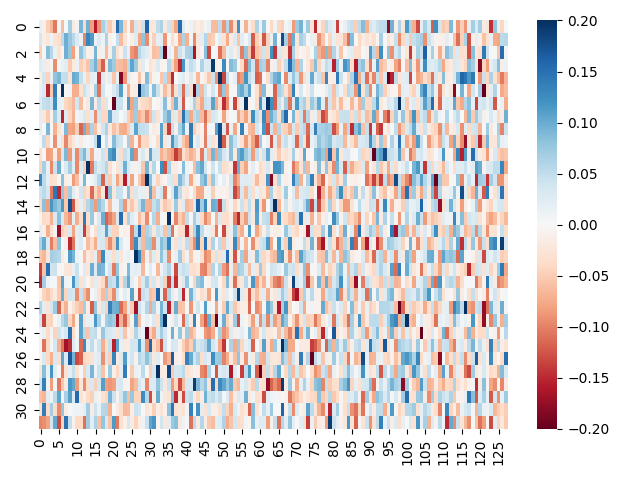}
         \caption{Without regularization}
         \label{fig:no-reg}
     \end{subfigure}
     \hfill
     \begin{subfigure}[b]{0.52\textwidth}
         \centering
         \includegraphics[width=\textwidth]{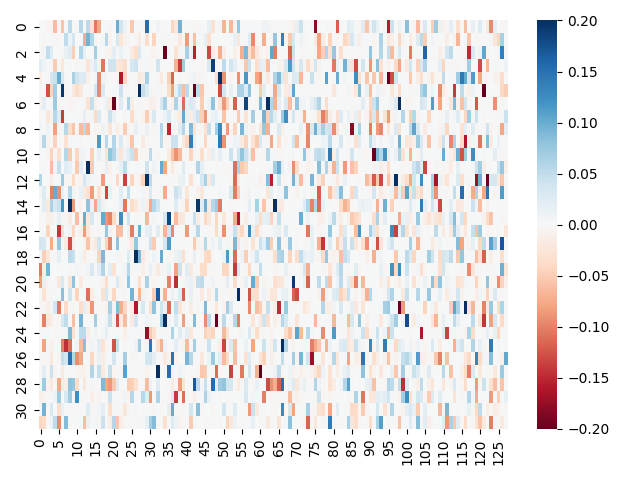}
         \caption{With group lasso regularization parameter $\lambda=0.01$ and $\ell_1$ regularization parameter $\gamma=0.01$.}
         \label{fig:with-reg}
     \end{subfigure}
    \caption{Intervened vector values across LLAMA3-8B attention heads (row-wise, from 1-32). Adding regularization promotes sparsity with the intervened values and desirable properties following previous empirical observations.}
    \label{fig:effect-reg-intervention}
\end{figure}

\section{Datasets}\label{sec:datasets}
The task descriptions of the rule understanding experiments are as follows:
\begin{itemize}[leftmargin=5mm]
    \item \textbf{Antonym}: Given an English word, generate an English word with the opposite meaning.
    \item \textbf{Synonym}: Given an English word, generate an English word with the same meaning.
    \item \textbf{English-French}: Given an English word, generate the equivalent word in French.
    \item \textbf{English-German}: Given an English word, generate the equivalent word in German.
\end{itemize}

\section{Prompts to measure GPT-Eval metric}

In this section, we provide the prompts to ask GPT-4 to confirm the input-prediction pair for each dataset in the Rule Understanding task.
\begin{itemize}
    \item \textbf{Antonym}: Answer 0 if what I say is wrong and 1 if it is correct. ``input'' is an antonym of ``prediction''.
    \item \textbf{Synonym}: Answer 0 if what I say is wrong and 1 if it is correct. ``input'' is a synonym of ``prediction''.
    \item \textbf{English-French}: Answer 0 if what I say is wrong and 1 if it is correct. ``input'' translated to French is ``prediction''.
    \item \textbf{English-German}: Answer 0 if what I say is wrong and 1 if it is correct. ``input'' translated to German is ``prediction''.
\end{itemize}
It is worth noting that ``input'' and ``prediction'' are placeholders and should be replaced with the actual input-prediction pair.

\end{document}